%% file: main_icme.tex
\documentclass[conference]{IEEEtran}
\IEEEoverridecommandlockouts
\usepackage{cite}
\usepackage{amsmath,amssymb,amsfonts}
\usepackage{algorithmic}
\usepackage{graphicx}
\usepackage{textcomp}
\usepackage{xcolor}
\def\BibTeX{{\rm B\kern-.05em{\sc i\kern-.025em b}\kern-.08em
    T\kern-.1667em\lower.7ex\hbox{E}\kern-.125emX}}

\usepackage{adjustbox}
\usepackage{url}
\usepackage{booktabs}
\usepackage{multirow}
\usepackage{hyperref}

\begin{document}

\title{Learnable Query Aggregation with KV Routing for Cross-view Geo-localisation}

\author{
    Hualin Ye\textsuperscript{1,*}, Bingxi Liu\textsuperscript{2,*}, \IEEEmembership{Graduate Student Member, IEEE}, Jixiang Du\textsuperscript{1,\dag}, \\ 
    Yu Qin\textsuperscript{1}, Ziyi Chen\textsuperscript{1}, \IEEEmembership{Senior Member, IEEE}, 
    and Hong Zhang\textsuperscript{2}, \IEEEmembership{Life Fellow, IEEE}
}

\maketitle
\begingroup
    \renewcommand{\thefootnote}{}
    
    \footnotetext{
        \scriptsize 
        \noindent\rule{1in}{0.4pt}\par\vspace{2pt} 
        \textsuperscript{1}Huaqiao University, Xiamen, China.
    }
    
    \footnotetext{
        \scriptsize
        \textsuperscript{2}Southern University of Science and Technology, Shenzhen, China.
    }
    \footnotetext{
        \scriptsize
        \textsuperscript{\dag}Corresponding author.
    }
    \footnotetext{
        \scriptsize
        \textsuperscript{*}Equal contribution.
    }
\endgroup

\input{sections/0_abstract}

\input{sections/1_introduction}

\input{sections/2_method}

\input{sections/3_experiments}

\input{sections/4_conclusion}




\bibliographystyle{IEEEbib}
\bibliography{ref}

\end{document}

%% file: sections/0_abstract.tex
\begin{abstract}
Cross-view geo-localisation (CVGL) aims to estimate the geographic location of a query image by matching it with images from a large-scale database. However, the significant viewpoint discrepancies present considerable challenges for effective feature aggregation and alignment.
To address these challenges, we propose a novel CVGL system that incorporates three key improvements. Firstly, we leverage the DINOv2 backbone with a convolution adapter fine-tuning to enhance model adaptability to cross-view variations. Secondly, we propose a multi-scale channel reallocation module to strengthen the diversity and stability of spatial representations. Finally, we propose an improved aggregation module that integrates a Mixture-of-Experts (MoE) routing into the feature aggregation process. Specifically, the module dynamically selects expert subspaces for the keys and values in a cross-attention framework, enabling adaptive processing of heterogeneous input domains. Extensive experiments on the University-1652 and SUES-200 datasets demonstrate that our method achieves competitive performance with fewer trained parameters.
\end{abstract}



%% file: sections/1_introduction.tex
\section{Introduction}
\label{sec:intro}

Cross-view geo-localisation (CVGL) refers to estimating the geographic location of a query image by comparing it with geo-referenced images from a pre-collected database based on visual similarity~\cite{GEOCAPSNET, wang2021each}. As a novel positioning paradigm that does not rely on satellite signals, CVGL compensates for the deficiencies of traditional Global Navigation Satellite System (GNSS) positioning, which is often compromised in complex real-world environments, such as forests and urban canyons, by factors like tree occlusion and electromagnetic interference that lead to signal attenuation. Consequently, CVGL has been widely employed in several critical domains, including unmanned aerial vehicle (UAV) localisation, and military reconnaissance~\cite{SMDT,huang2024cv}.

Early CVGL methods primarily relied on hand-crafted features for localisation~\cite{lin2013cross}. However, due to challenges such as cross-viewpoint variations and occlusions, hand-crafted features often fail to maintain stable discriminative capability, leading to suboptimal localisation performance.

In recent years, learning-based methods have demonstrated remarkable capabilities in feature extraction and generalisation across a wide range of computer vision tasks. With the continuous development of convolutional neural networks, several studies have incorporated spatial cues or keypoint-based information to extract fine-grained features, thereby enhancing geo-localisation performance to some extent~\cite{LPN, keypoint}. Recent, models such as ConvNeXt and DINO are capable of learning rich semantic and structural representations, achieving promising results in CVGL tasks~\cite{huang2024cv, deuser2023sample4geo}. However, these studies require the fine-tuning of \textit{a large number of parameters} when transferring pre-trained models to CVGL tasks. Inspired by Parameter-Efficient Fine-Tuning (PEFT), we \textit{introduce} convolutional adaptation tuning\cite{jie2022convolutional} to CVGL, which significantly reduces the number of trainable parameters while effectively mitigating catastrophic forgetting. Additionally, we also \textit{propose} a multi-scale channel reallocation (MSCR) to the pre-trained model to bridge the gap between the visual foundation models and remote sensing tasks.

\begin{figure}[t]
\centering
\includegraphics[width=\linewidth]{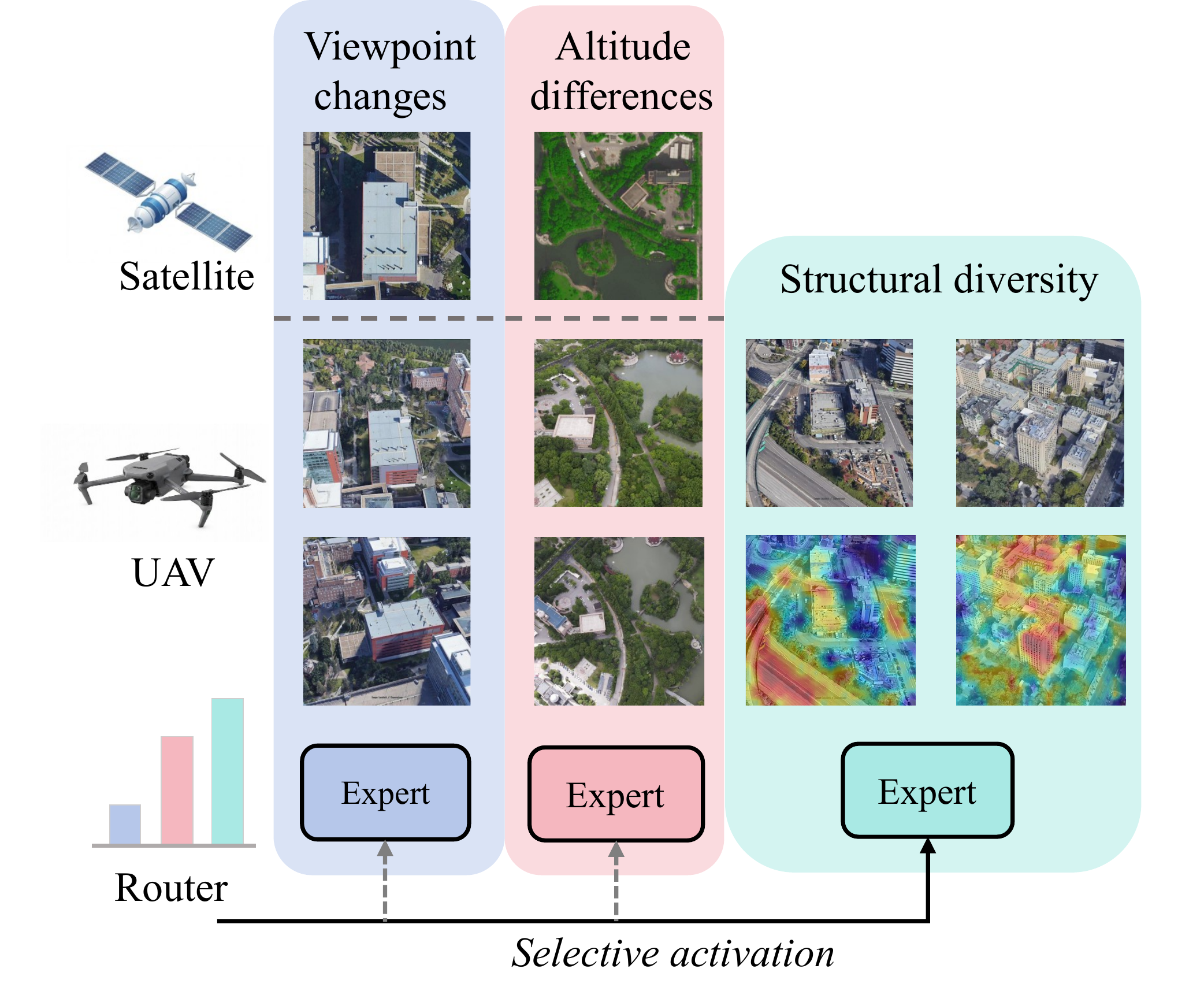}
\caption{Each column presents images from the same location. The top row shows satellite imagery, while the second and third rows depict UAV-captured views. UAV images from a single location often exhibit substantial variations in scale and viewpoint, whereas images from different locations tend to contain highly diverse discriminative structures (e.g., roads and buildings). By leveraging specialised expert models, the MoE-enhanced aggregation layer selectively activates the most suitable experts, thereby improving performance on these challenging cases.}
\label{fig:tier}
\end{figure}

\begin{figure*}[ht]
\centering
\includegraphics[width=\textwidth]{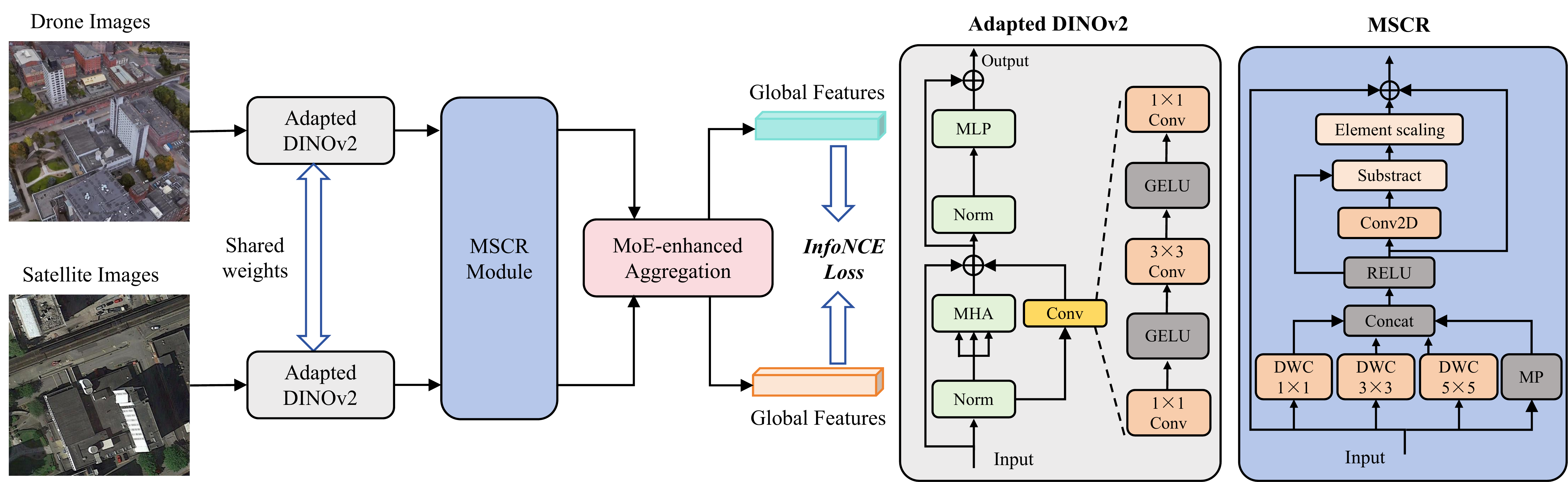}
\caption{The proposed system is illustrated above, comprising three contributions: (1) convolution-based adaptation, (2) the MSCR module, and (3) an MoE-enhanced aggregator. The right-hand side provides architectural details of the first two components, while the structure of the aggregator is presented on the following page.}
\label{fig:framework}
\end{figure*}

In CVGL, another core problem is how to aggregate the deep features extracted from images into a compact, discriminative global descriptor. Prior work has typically adopted aggregation strategies from Visual Geo-localisation, such as GeM, NetVLAD \cite{netvlad}, and SALAD \cite{SALAD}. More recently, Bag-of-Queries (BoQ) \cite{BoQ} introduced a set of global learnable queries $Q$ that are input‑independent parameters. During inference, they utilise cross-attention to probe local features and aggregate them into a consistent global representation. This design avoids dynamically deriving queries from each input and therefore offers more stable global descriptors across diverse environments, such as viewpoint changes or weather differences. 

Motivated by this, we \textit{propose} to go one step further: we design an MoE‑enhanced aggregation module that expands the expressivity of the keys $K$ and values $V$. Specifically, by employing a Mixture-of-Experts (MoE) style mechanism, our module utilises a routing mechanism to dynamically select a small subset of "expert" subspaces for $K$ and $V$ based on the features. As shown in Fig.~\ref{fig:tier}, this KV‑routing design allows the network to adaptively pick different experts to process different types of inputs, thereby capturing domain‑specific cues more effectively. As a result, our aggregator not only inherits BoQ’s ability to provide a compact, input‑agnostic global representation but also gains robustness and flexibility when tackling the domain shifts inherent in CVGL tasks. \textbf{The contributions of this work are summarised as follows:}
\begin{itemize}
    \item We introduce a DINOv2-based backbone in combination with a convolutional adapter fine-tuning approach, which enhances the model's adaptability for cross-view geo-localisation tasks.
    \item We propose a novel Multi-scale Channel Reallocation module, which improves the spatial structural perception of geo-localisation features across varying scales.
    \item We propose a MoE-enhanced aggregation that dynamically adapts to domain-specific cues, offering superior flexibility and robustness in handling domain shifts within CVGL tasks.
    \item Our method demonstrates competitive performance on two benchmark datasets, and the ablation studies validate the effectiveness of the aforementioned contributions.
\end{itemize}

%% file: sections/2_method.tex
\section{Methodology}
\label{sec:methodology}



The framework of the proposed method is illustrated in Fig.~\ref{fig:framework}. We adopt DINOv2 as the backbone network owing to its strong general-purpose representation capability for extracting visual features. To transfer its generalisation ability for cross-view localisation, we employ a convolution-based adapter for fine-tuning in Sec.~\ref{sec:2.1}, and introduce a multi-scale channel reallocation module in Sec.~\ref{sec:2.2}. In Sec.~\ref{sec:2.3}, we present an improved feature aggregation that leverages a mixture-of-experts mechanism to provide optimal inputs to the Transformer-based aggregator via the Key and Value pathways.

\subsection{Cross-view Fine-Tuning based on Convolutional Adapters}
\label{sec:2.1}

The pre-training data of DINOv2 consist primarily of ground-level photographs, which differ substantially from UAV or satellite imagery in terms of viewpoint, scale, and textural characteristics. Consequently, directly applying DINOv2 to CVGL tasks typically results in \textit{suboptimal} performance. In addition, \textit{full} fine-tuning of such a large-scale foundation model incurs considerable computational overhead.

To address this limitation, we introduce a lightweight adapter-based tuning strategy. Specifically, we insert a convolutional adapter~\cite{jie2022convolutional} into each Transformer block $\mathcal{T}$ of DINOv2 and update \textit{only} the adapter parameters while keeping the backbone frozen. This adapter follows a bottleneck-style design consisting of a $1{\times}1$ convolution, a $3{\times}3$ convolution, and another $1{\times}1$ convolution, each separated by GELU activations. Formally, for an input feature map $\mathbf{z}$, the adapter is defined as
\begin{equation}
    \mathcal{A}_l(\mathbf{z}) 
    = W^{(3)}_l \,
      \sigma\!\left(
      \mathrm{Conv}_{3 \times 3}\!\left(
      \sigma\!\left(
      W^{(1)}_l \mathbf{z}
      \right)\right)\right),
\end{equation}
where $W^{(1)}_l$ and $W^{(3)}_l$ denote the $1{\times}1$ convolutions, 
$\mathrm{Conv}_{3\times 3}$ is the spatial convolution, and $\sigma(\cdot)$ is the GELU activation.

The adapter-enhanced Transformer block is thus written as
\begin{equation}
    \mathbf{z}_{l+1}
    = \mathcal{T}_l(\mathbf{z}_l)
    + \mathcal{A}_l(\mathbf{z}_l),
\end{equation}
with only the adapter parameters being trainable.

In contrast to language-oriented adapters (LoRA), the convolutional adapter provides a vision-aligned inductive bias that captures local geometric patterns, thereby alleviating the severe viewpoint discrepancies inherent in cross-view matching.




\subsection{Multi-scale Channel Reallocation}
\label{sec:2.2}

While adapters help alleviate domain shift, cross-view geo-localisation in remote sensing still confronts substantially larger scale discrepancies than ground-to-ground localisation. To mitigate this mismatch, we introduce the Multi-scale Channel Reallocation (MSCR) module, which is designed to strengthen the model’s capacity to reconcile geographic structures across varying spatial scales.

\subsubsection{Multi-scale Feature Extraction} The MSCR module employs a multi-branch depthwise feature extraction mechanism to obtain diverse spatial cues. Specifically, an input feature map $X$ is simultaneously processed by depthwise separable convolutions with kernel sizes $1{\times}1$, $3{\times}3$, and $5{\times}5$, together with a max-pooling branch:
\begin{equation}
    X_\text{kernel} = \mathrm{DWConv}_{\text{kernel}\times\text{kernel}}(X), \quad
    X_{\mathrm{mp}} = \mathrm{MaxPool}(X).
\end{equation}

These outputs are concatenated to produce a multi-scale feature representation:
\begin{equation}
    X_{\mathrm{ms}} = \mathrm{Concat}(X_{1},\, X_{3},\, X_{5},\, X_{\mathrm{mp}}).
\end{equation}

A GELU activation and a dropout layer are then applied for non-linear transformation and regularisation:
\begin{equation}
    X_{\mathrm{loc}} = \mathrm{Dropout}(\mathrm{GELU}(X_{\mathrm{ms}})).
\end{equation}

\subsubsection{Residual Signal Modulation} To suppress redundant responses and enhance salient geographic cues, MSCR incorporates a residual signal modulation mechanism\cite{MogaNet}. A $1{\times}1$ convolution compresses the local contextual features into a single-channel residual signal:
\begin{equation}
    R = \mathrm{Conv}_{1\times1}(X_{\mathrm{loc}}),
\end{equation}
which models the redundancy distribution in the feature space.

The residual-enhanced feature is obtained via subtraction:
\begin{equation}
    X_{\mathrm{res}} = X_{\mathrm{loc}} - R.
\end{equation}

To adaptively regulate the strength of enhancement, a learnable element-wise scaling parameter $\sigma$ is introduced:
\begin{equation}
    X_{\mathrm{adj}} = X_{\mathrm{loc}} + \sigma \odot X_{\mathrm{res}},
\end{equation}
where $\odot$ denotes element-wise multiplication.

\subsubsection{Final Output} Finally, a residual connection integrates the enhanced representation with the module's original input:
\begin{equation}
    Y = X + X_{\mathrm{adj}},
\end{equation}
producing a multi-scale- aware feature representation that is well-suited for cross-view image retrieving.

\begin{figure}[!htbp]
\centering
\includegraphics[width=0.8\linewidth]{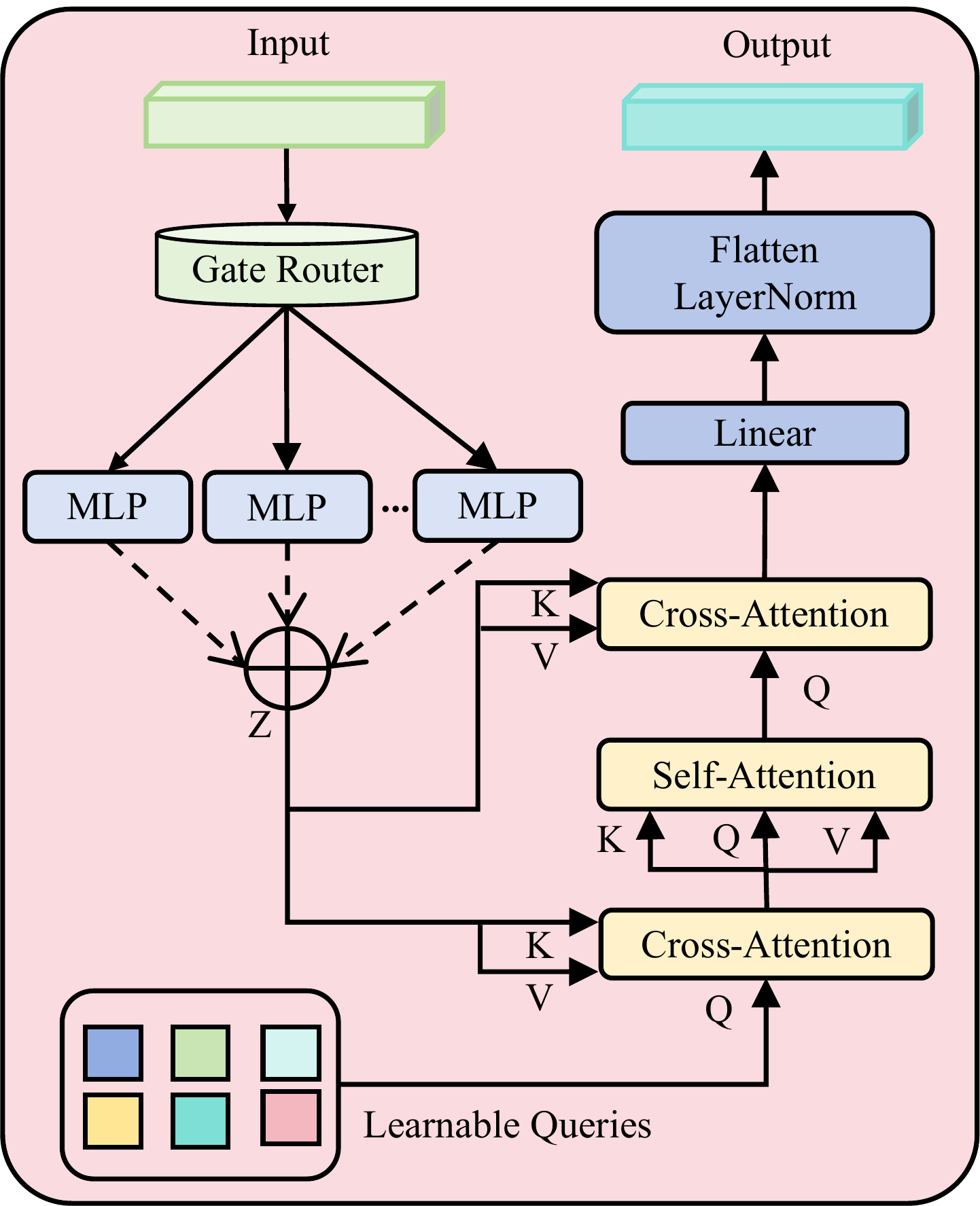}
\caption{Detailed illustration of the BoQ module with KV routing.}
\label{fig:moe_boq}
\end{figure}

\subsection{MoE-enhanced Aggregation}
\label{sec:2.3}

Having obtained the feature tokens, a central challenge in visual geo-localisation is their effective aggregation into a compact yet highly discriminative global descriptor. Inspired by the Bag-of-Queries (BoQ)\cite{BoQ} architecture, which improves domain generalisation by treating the query vectors $Q$ as learnable and data-independent parameters, we introduce an MoE-enhanced aggregation module. This module employs a Mixture-of-Experts design to enhance the representational capacity of the keys $K$ and values $V$, allowing the network to adaptively encode domain-specific cues that emerge in cross-view retrieval.

\subsubsection{BoQ-like Aggregation}

Following BoQ, we maintain a fixed set of learnable query vectors
\begin{equation}
Q = {q_1, q_2, \ldots, q_K} \in \mathbb{R}^{K \times D},
\end{equation}
which act as trainable parameters independent of the input tokens. Unlike conventional self-attention, where queries, keys, and values stem from the same feature map, the BoQ formulation explicitly decouples the queries, allowing them to function as content-agnostic filters. These queries are progressively refined through \textit{a three-stage attention} pipeline applied to the expert-enhanced input tokens.

At each stage, information is exchanged between the learnable queries and the feature tokens through standard multi-head attention:
\begin{equation}
\mathrm{MultiHeadAttn}(Q, K, V)
= \operatorname{softmax}\left(\frac{Q K^{\top}}{\sqrt{d}}\right) V,
\end{equation}
enabling the queries to identify and aggregate the most relevant tokens.
This design retains the essential property of BoQ: the global descriptor is produced not by pooling over feature tokens, but by iteratively querying them, resulting in a flexible and highly expressive representation.

\subsubsection{Mixture-of-Experts for Keys and Values}

While BoQ introduces learnability into the queries, the representation of the keys and values remains restricted to a single shared projection, limiting their ability to capture domain-specific patterns. Our MoE-enhanced design routes feature tokens through multiple experts, each modelling different semantic or appearance variations commonly present in cross-view geo-localisation, such as viewpoint changes, altitude differences, structural diversity, and environmental conditions.

Given the input token sequence \(X \in \mathbb{R}^{N \times D}\), we introduce \(E\) experts,
\[
\mathcal{E} = \{f_{1}, f_{2}, \ldots, f_{E}\},
\]
where each expert \(f_{e}\) is implemented as a lightweight feed-forward module that generates expert-specific keys and values:
\begin{equation}
K_{e}, V_{e} = f_{e}(X).
\end{equation}
A gating network \(G(\cdot)\) produces a sparse routing distribution:
\begin{equation}
g = G(X) \in \mathbb{R}^{N \times E},
\end{equation}
where a top-1 gating strategy ensures computational efficiency.

The final expert-enhanced keys and values are obtained via a sparse mixture:
\begin{align}
K = \sum_{e=1}^{E} g_{e} \odot K_{e},
V = \sum_{e=1}^{E} g_{e} \odot V_{e}.
\end{align}

The resulting MoE-enhanced \(K\) and \(V\) are subsequently fed into the BoQ-like aggregation pipeline, allowing the learnable queries to extract information from a richer and more specialised token space. This results in a global descriptor with significantly enhanced discriminative ability in cross-view retrieval tasks.

\subsection{Loss Function}

Inspired by the latest developments in cross-view image retrieving, the proposed model is trained using the symmetric InfoNCE loss\cite{deuser2023sample4geo}. This objective leverages all negative pairs within each batch, which mitigates the randomness of negative sampling and enhances the model’s ability to distinguish between different negative instances. This design ultimately improves scalability and generalisation. The loss function is formulated as follows:
\begin{equation}
    \mathcal{L}(q, R)_{\text {InfoNCE }}=-\log \frac{\left.\exp \left(q \cdot r_{+} / \tau\right)\right)}{\left.\sum_{i=0}^{R} \exp \left(q \cdot r_{i} / \tau\right)\right)},
\end{equation}
which \(q\) denotes the feature encoding of the query image, and \(R\) represents the set of feature encodings of all reference images within a batch. This set contains only one positive sample \(r_{+}\) that matches \(q\), while the rest are negative samples. The temperature coefficient \(\tau\) is a hyperparameter that can be either learnable or fixed to a constant value.

%% file: sections/3_experiments.tex
\section{Experiment}

\subsection{Experimental Settings}

\input{tables/datasets}

\subsubsection{Dataset} We conduct experiments on two cross-view geo-localisation benchmarks: University-1652 and SUES-200. The University-1652 dataset comprises imagery from 1,652 buildings across 72 universities worldwide, containing both drone-view photographs and satellite orthophotos \cite{zheng2020university}. The SUES-200 dataset similarly provides UAV-view and satellite-view imagery collected around ShanghaiTech University, covering a diverse set of urban and suburban scenes. Moreover, UAV imagery in SUES-200 is captured at multiple flight altitudes (150 m, 200 m, 250 m, and 300 m) \cite{zhu2023sues}. The number of queries and gallery samples for both datasets is summarised in Tab.~\ref{tab:dataset}.

\subsubsection{Implementation Details} We implement our method in PyTorch, adopting DINOv2 with a ViT-B/14 backbone. Input images are uniformly resized to 322 $\times$ 322. The initial learning rate is set to 0.005 and scheduled using cosine annealing. We use the Adam optimiser with a batch size of 24, and train the network for 40 epochs in total.


\subsubsection{Evaluation Metrics} In cross-view geo-localisation, models are commonly evaluated using R@K and Average Precision (AP). R@K denotes the proportion of query samples for which at least one correct match appears among the top-$K$ retrieved results. AP is computed as the area under the precision–recall curve, jointly reflecting retrieval precision and recall.

\subsection{Comparison to Other Methods}

To demonstrate the effectiveness of the proposed method, comparative experiments were conducted against SOTAs on the University-1652 and SUES-200 datasets.

\input{tables/s200}

\subsubsection{Results on the SUES-200 dataset} To comprehensively evaluate the generality and superiority of the proposed method, experiments were conducted on the SUES-200 dataset. As shown in Tab. \ref{tab:sues200}, for the drone-to-satellite task, the R@1 scores at UAV altitudes of 150 m, 200 m, 250 m, and 300 m reached 94.7, 97.93, 98.73, and 98.85, respectively, while the corresponding AP values were 95.70, 98.34, 98.73, and 98.85. In the satellite-to-drone task, while the R@1 scores are near saturation (98.75) across all altitudes, our AP values at the four altitudes were 95.61, 97.7, 98.47, and 98.91, respectively, consistently outperforming other methods. These results demonstrate the robustness of our approach across different flight altitudes, maintaining competitive performance on multiple metrics.

\input{tables/u1652}

\subsubsection{Results on the University-1652 dataset} The performance of the proposed method was also compared with other existing methods, as summarised in Tab. \ref{tab:performance}. For the satellite-to-drone task, our method achieved an R@1 of 96.72 and an AP of 93.57. In the drone-to-satellite task, the R@1 and AP achieved 94.41 and 95.40, respectively, outperforming other current methods. It is noteworthy that our model contains \textit{only 10 M} parameters, which is significantly lower than the mainstream methods listed in the table, accounting for merely 30\% of the MEAN parameter count.

\subsection{Ablation Study}


\input{tables/ablation_ft}

\input{tables/ablation_moe}

\input{tables/ablation_expert}

\subsection{Visualizations}

\begin{figure}[h]
\centering
\includegraphics[width=\linewidth]{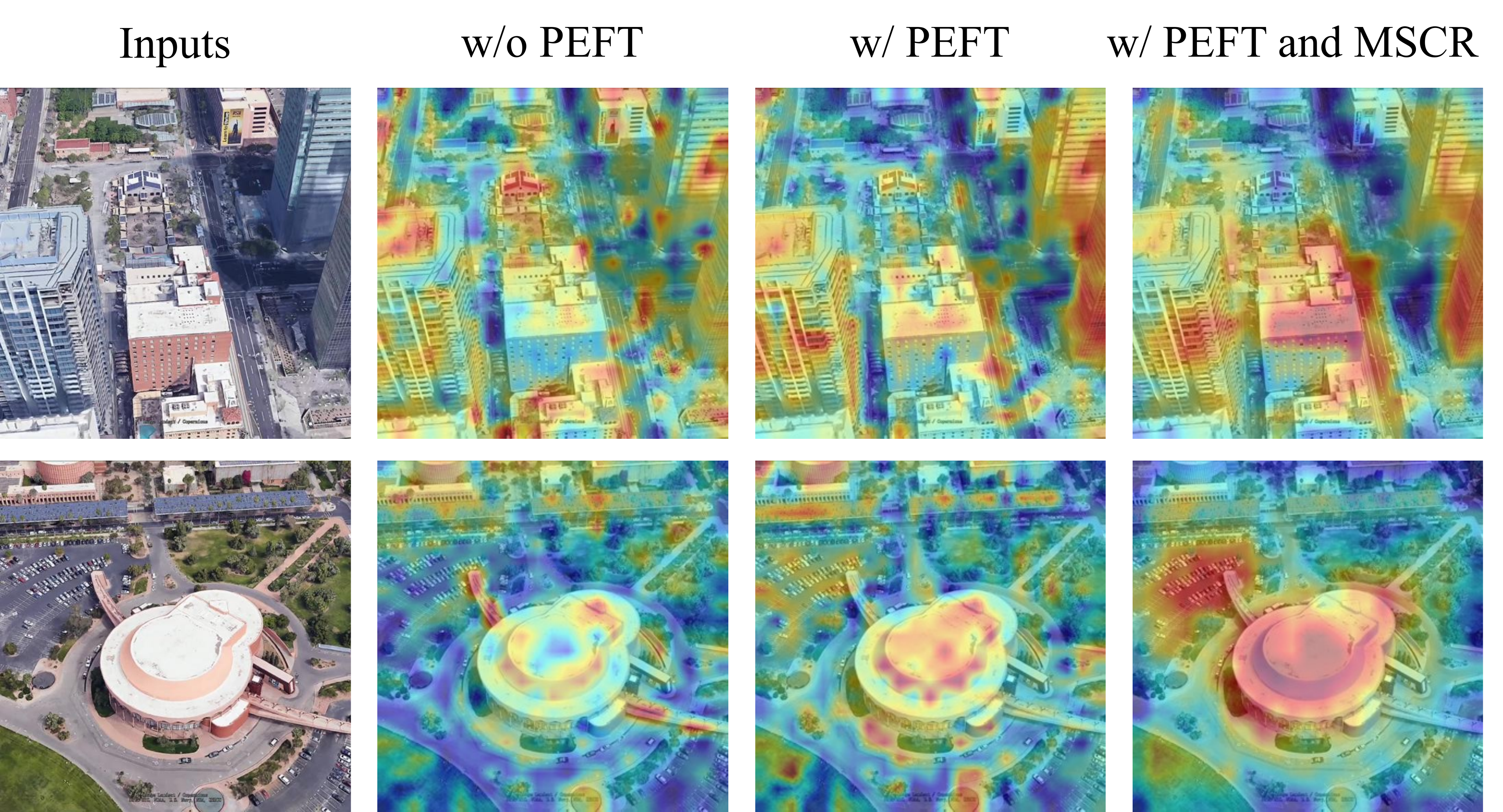}
\caption{Feature map visualisations for the ablation study from two scenes. From left to right: the original input image, w/o PEFT (fully frozen DINOv2), w/ PEFT (DINOv2 fine-tuned via Convolution), and our method (full model incorporating both Convolution fine-tuning and the MSCR module).}
\label{fig:vis}
\end{figure}

The visualisations in Fig. \ref{fig:vis} compare the attention responses of the model under three configurations: without PEFT, with PEFT, and with both PEFT and the proposed MSCR module. Without PEFT, the model exhibits scattered and noisy activation patterns, failing to consistently focus on semantically meaningful structures. Introducing PEFT leads to noticeably sharper and more coherent responses, indicating improved feature adaptation to cross-view discrepancies. When MSCR is further incorporated, the activation maps become more concentrated around key architectural and geometric cues, demonstrating enhanced multi-scale awareness. These results confirm that PEFT and MSCR further strengthens the model’s ability for robust cross-view matching.

%% file: tables/datasets.tex
\begin{table}[!t]
\centering
\caption{Statistics of the University-1652 and SUES-200 Datasets for Two Different CVGL Tasks.}
\label{tab:dataset}
\begin{tabular}{l c c c c}
\toprule
\multirow{2}{*}{\textbf{Dataset}} & \multicolumn{2}{c}{\textbf{Drone$\rightarrow$Satellite}} & \multicolumn{2}{c}{\textbf{Satellite$\rightarrow$Drone}} \\
\cmidrule(lr){2-3} \cmidrule(lr){4-5}
 & \textbf{Query} & \textbf{Gallery} & \textbf{Query} & \textbf{Gallery} \\
\midrule
University-1652 & 37855 & 951 & 701 & 51355 \\
SUES-200        & 16k / 4k & 200 & 80 & 40k / 10k \\
\bottomrule
\end{tabular}
\end{table}

%% file: tables/s200.tex
\begin{table*}[!t]
\caption{Comparison of Different Methods on the SUES-200 Dataset (Drone$\rightarrow$Satellite and Satellite$\rightarrow$Drone)}
\label{tab:sues200}
\centering
\begin{adjustbox}{width=\linewidth}
\begin{tabular}{l|cccccccc|cccccccc}
\hline
\multirow{3}{*}{Methods} 
& \multicolumn{8}{c|}{Drone $\rightarrow$ Satellite} 
& \multicolumn{8}{c}{Satellite $\rightarrow$ Drone} \\
\cline{2-17}
& \multicolumn{2}{c}{150m} & \multicolumn{2}{c}{200m} & \multicolumn{2}{c}{250m} & \multicolumn{2}{c|}{300m}
& \multicolumn{2}{c}{150m} & \multicolumn{2}{c}{200m} & \multicolumn{2}{c}{250m} & \multicolumn{2}{c}{300m} \\
\cline{2-17}
& R@1 & AP & R@1 & AP & R@1 & AP & R@1 & AP 
& R@1 & AP & R@1 & AP & R@1 & AP & R@1 & AP \\
\hline
Sample4geo\cite{deuser2023sample4geo}
& 92.60 & 94.00 & 97.38 & 97.81 & 98.28 & 98.64 & \textbf{99.18} & \textbf{99.36}
& 97.50 & 93.63 & 98.75 & 96.70 & 98.75 & 98.28 & 98.75 & 98.05 \\

FDER\cite{ge2024FDER}
& 85.30 & 87.58 & 93.23 & 94.66 & 94.47 & 97.28 & 97.50 & 88.09
& 93.75 & 86.93 & 97.75 & 93.12 & 98.75 & 96.81 & 98.75 & 97.20 \\

MFJR\cite{ge2024MFJR}
& 88.95 & 91.05 & 93.60 & 94.72 & 95.42 & 96.28 & 97.45 & 97.84
& 95.00 & 89.31 & 96.25 & 94.72 & 94.69 & 96.92 & 98.75 & 97.87 \\

SeGCN\cite{liu2024segcn}
& 90.80 & 92.32 & 91.93 & 93.41 & 92.53 & 93.90 & 93.33 & 94.61
& 93.75 & 92.45 & 95.00 & 93.65 & 96.25 & 94.39 & 97.50 & 94.55 \\

CCR\cite{du2024ccr}
& 87.08 & 95.55 & 93.57 & 94.90 & 95.42 & 96.28 & 96.82 & 97.39
& 92.50 & 88.54 & 97.50 & 95.22 & 97.50 & 97.10 & 97.50 & 97.49 \\

DAC\cite{xia2024DAC}
& \textbf{96.80} & \textbf{97.54} & 97.48 & 97.97 & 98.20 & 98.62 & 97.58 & 98.14
& 97.50 & 94.06 & 98.75 & 96.66 & 98.75 & 98.09 & 98.75 & 97.87 \\

\textbf{Ours}
& 94.70 & 95.70 & \textbf{97.93} & \textbf{98.34} & \textbf{98.73} & \textbf{99.02} & 98.85 & 99.11
& \textbf{98.75} & \textbf{95.61} & 98.75 & \textbf{97.70} & 98.75 & \textbf{98.47} & 98.75 & \textbf{98.91} \\
\hline
\end{tabular}
\end{adjustbox}
\end{table*}

%% file: tables/u1652.tex
\begin{table}[!t]
 \caption{Comparison of Different Cross-view Geo-Localization Methods on University-1652}
    \label{tab:performance}
    \centering
    \begin{adjustbox}{width=\linewidth}
        \begin{tabular}{l c c c c c}
        \toprule
       \multirow{2}{*}{Methods} & \multirow{2}{*}{Trained} & \multicolumn{2}{c}{Satellite To Drone} & \multicolumn{2}{c}{Drone To Satellite} \\
        \cmidrule(lr){3-4} \cmidrule(lr){5-6}
         & Params & R@1 & AP & R@1 & AP \\
        \midrule
        Sample4ego (\textit{ICCV'23})\cite{deuser2023sample4geo} & 88.60 & 95.14 & 91.39 & 92.65 & 93.81 \\
        MCCG (\textit{TCSVT'23})\cite{shen2023mccg} & 56.65 & 94.30 & 89.39 & 89.64 & 91.32 \\
        FDER (\textit{TGRS'24})\cite{ge2024FDER} & 97.13 & 95.58 & 92.17 & 92.79 & 93.91 \\
        CCR (\textit{TCSVT'24})\cite{du2024ccr} & 156.57 & 95.15 & 91.80 & 92.54 & 93.78 \\
        SDPL  (\textit{TCSVT'24})\cite{chen2024sdpl} & 42.56 & 93.58 & 89.45 & 90.16 & 91.40 \\
        MEAN (\textit{TGRS'25})\cite{chen2025MEAN} & 36.50 & 96.01 & 92.08 & 93.55 & 94.53 \\
        \textbf{Ours} & \textbf{10.61} & \textbf{96.72} & \textbf{93.57} & \textbf{94.41} & \textbf{95.40} \\
        \bottomrule
    \end{tabular}
    \end{adjustbox}

\end{table}

%% file: tables/ablation_ft.tex
\begin{table}[t]
\caption{Ablation study on PEFT-based methods and the plugin module.}
\label{ablation_ft}
\centering
\begin{tabular}{l c c c c} 
    \toprule
    \multirow{2}{*}{Method} & \multicolumn{2}{c}{Satellite to Drone} & \multicolumn{2}{c}{Drone to Satellite} \\
    \cmidrule(lr){2-3} \cmidrule(lr){4-5}
     & R@1 & AP & R@1 & AP \\
    \midrule
   1: LoRA & 95.29 & 91.56 & 93.87 & 95.01 \\ 
   2: Conv-Adapter & 95.57 & 91.39 & 94.28 & 95.17 \\
   2 + MSCR & \textbf{96.72} & \textbf{93.57} & \textbf{94.41} & \textbf{95.40} \\ 
    \bottomrule
\end{tabular}
\end{table}

\subsubsection{Ablation on Module Contributions} Tab.~\ref{ablation_ft} reports the ablation study conducted on different PEFT-based fine-tuning strategies and the proposed plugin module. Compared with LoRA~\cite{LoRA}, the Conv-Adapter achieves slightly better performance across both transfer directions, indicating that convolutional adaptation provides a more suitable inductive bias for handling cross-view discrepancies. Incorporating the MSCR module yields a further and consistent improvement in both settings. This demonstrates that MSCR effectively enhances the model’s ability to capture discriminative structures under significant viewpoint and appearance variations.

%% file: tables/ablation_moe.tex
\begin{table}[t]
\caption{Ablation experiments on the effect of introducing MoE.}
\label{ablation_moe}
\centering
\begin{tabular}{l c c c c} 
    \toprule
    \multirow{2}{*}{Method} & \multicolumn{2}{c}{Satellite to Drone} & \multicolumn{2}{c}{Drone to Satellite} \\
    \cmidrule(lr){2-3} \cmidrule(lr){4-5}
     & R@1 & AP & R@1 & AP \\
    \midrule
    BoQ & 95.86 & 93.31 & 93.38 & 94.48 \\ 
    BoQ w/ KV Rounting & \textbf{96.72} & \textbf{93.57} & \textbf{94.41} & \textbf{95.40} \\ 
    \bottomrule
\end{tabular}
\end{table}

\subsubsection{Ablation on the Impact of MoE.} Tab. \ref{ablation_moe} evaluates the impact of incorporating MoE into the BoQ module. While the baseline \cite{BoQ} already provides strong performance, introducing KV routing within BoQ yields consistent improvements across both retrieval directions. Specifically, the MoE-enhanced variant achieves higher R@1 and AP scores, demonstrating its ability to more effectively capture discriminative cross-view correspondences. These gains confirm that selectively activating specialised experts helps the model better adapt to diverse structural patterns and viewpoint variations present in satellite and drone imagery.

%% file: tables/ablation_expert.tex
\begin{table}[ht]
\caption{Ablation Experiments on the Effectiveness of Model Components}
\label{ablation_expert}
\centering
\begin{tabular}{c c c c c} 
    \toprule
    \multirow{2}{*}{$K$} & \multicolumn{2}{c}{Satellite to Drone} & \multicolumn{2}{c}{Drone to Satellite} \\
    \cmidrule(lr){2-3} \cmidrule(lr){4-5}
     & R@1 & AP & R@1 & AP \\
    \midrule
    1 & 96.14 & 93.17 & 93.66 & 94.80 \\ 
    3 & \textbf{96.72} & \textbf{93.57} & \textbf{94.41} & \textbf{95.40} \\ 
    5 & 96.29 & 93.28 & 94.03 & 95.12 \\
    \bottomrule
\end{tabular}
\end{table}

\subsubsection{Ablation on the number of Experts.} Tab. \ref{ablation_expert} investigates the influence of varying the number of experts $K$ within the MoE-enhanced aggregator. Using a single expert already offers competitive results, indicating that the aggregation mechanism is intrinsically effective. Increasing the number of experts to $K$=3 yields the best performance across all metrics in both transfer directions, suggesting that a moderate level of expert specialisation allows the model to better capture diverse appearance patterns and cross-view variations. However, further increasing the number of experts to $K$=5 leads to a slight decline in performance, likely due to over-fragmentation of the feature space and reduced expert utilisation efficiency. These observations highlight the importance of balancing expert diversity and capacity, with $K$=3 providing the most favourable trade-off.

%% file: sections/4_conclusion.tex
\section{Conclusion}

In this work, we presented an efficient and adaptable framework for CVGL that integrates convolutional adapter tuning, multi-scale channel reallocation, and an MoE-enhanced aggregation strategy. By leveraging the strong representation ability of DINOv2 while significantly reducing trainable parameters, our method effectively mitigates catastrophic forgetting and better aligns visual foundation models with remote-sensing scenarios. The proposed aggregation module further improves robustness to domain shifts by dynamically selecting expert sub-spaces to model diverse spatial cues. Extensive experiments on University-1652 and SUES-200 demonstrate that our approach achieves competitive performance and strong generalisation, highlighting its potential for real-world geo-localisation applications.